\documentclass[lettersize,journal,onecolumn]{IEEEtran} 
\usepackage{amsmath,amsfonts}                          
\usepackage{pseudocode}
\usepackage[boxruled, vlined]{algorithm2e}

\usepackage{array}                                     
\usepackage[caption=false,font=normalsize,labelfont=sf,textfont=sf]{subfig}
\usepackage{textcomp}
\usepackage{stfloats}
\usepackage{url}
\usepackage{verbatim}
\usepackage{graphicx}                                
\usepackage{cite}                                    
\usepackage{hyperref}                                
\usepackage{comment}                                 
\hypersetup{
    linkcolor=blue,
    filecolor=magenta,      
    urlcolor=blue,
    pdftitle={Overleaf Example},
    pdfpagemode=FullScreen,
    }
\usepackage{color}
\newcommand{\editb}[1]{{\color{black}\textbf{}#1}} 
\newcommand{\editr}[1]{{\color{black}\textbf{}#1}} 
\newcommand{\editg}[1]{{\color{black}\textbf{}#1}}  
\newcommand{\editm}[1]{{\color{black}\textbf{}#1}} 

\begin{document}

\title{Biased hypothesis formation from projection pursuit}



\author{
    John Patterson*$^{1}$, Chris Avery*$^{1}$, Tyler Grear $^{1,2}$ Donald J. Jacobs $\dagger$  $^{2,3}$ \\
    $^{1}$ Department of Bioinformatics and Genomics, University of North Carolina Charlotte, \\
    $^{2}$ Department of Physics and Optical Science, University of North Carolina Charlotte, \\
    $^{3}$ Affiliate faculty of the UNC Charlotte School of Data Science \\
    * equal contributions to work \\  
    $\dagger $ corresponding author: djacobs1@uncc.edu
}

\markboth{Advances in Artificial Intelligence and Machine Learning,~Vol.~?, Iss.~?, December~2021}%
{Shell \MakeLowercase{\textit{et al.}}: A Sample Article Using IEEEtran.cls for IEEE Journals} 

\date{December 2021}
\maketitle
    
\begin{abstract}
The \editm{effect of bias} on hypothesis formation is characterized for an automated data-driven projection pursuit neural network to extract and select features for binary classification of data streams. This intelligent exploratory process \editm{partitions a complete vector state space into disjoint subspaces to create} working hypotheses quantified by similarities and differences observed between two groups of labeled data streams. \editm{Data streams are typically time sequenced, and may exhibit complex spatio-temporal patterns. For example, given atomic trajectories from molecular dynamics simulation, the machine's task is to quantify dynamical mechanisms that promote function by comparing protein mutants, some known to function while others are nonfunctional. Utilizing} synthetic two-dimensional molecules that mimic the dynamics of functional and nonfunctional proteins, biases are identified and controlled in both the machine learning model and selected training data under different contexts. The refinement of a working \editm{hypothesis} converges to a statistically robust multivariate perception of the data based on a context-dependent perspective. \editm{Including} diverse perspectives \editm{during data exploration enhances} interpretability \editm{of the multivariate characterization of similarities and differences.}

\end{abstract}

\section{Introduction}
As machine learning (ML) becomes an integral part of human life, recent concerns about biases in ML predictions have arisen \cite{cummings2021subjectivity, gianfrancesco2018potential}. Analogous to self-directed category learning by humans \cite{Markant2018}, biases effect hypothesis formation during data exploration using modulated perspectives within a neural network (NN). Generally, NN models yield non-unique optimal perceptron weights \cite{sussman1992uniqueness}; furthermore, these weights depend on implementation details \cite{hochreiter1998the, sun2020a} that contribute to biases. Unfortunately, results from a NN are often difficult to interpret; by extension, the underlying biases are difficult to characterize. In contrast, biases can be effectively controlled with an objective function within projection pursuit (PP) during the exploration of high-dimensional data \cite{kruskal1969toward, freidman1974a, huber1985projection, hyvarinen2000independent, martinez2001pca, lee2005projection, Cook2021, espezua2015a, grear2021molecular}. Moreover, PP is robust against statistical estimation errors \cite{friedman1997on}, and perception is interpretable by linear projection operators that govern dimension reduction. 

A new intelligent ML paradigm known as supervised projective learning with orthogonal completeness (SPLOC) is employed here. As an automated PP framework optimized by a recurrent NN, SPLOC performs a data-driven process for binary discriminant analysis of data streams. For example, molecular dynamics simulations containing dozens of molecules, each involving thousands of degrees of freedom (df) comprising tens of thousands sampled conformations, are feasible to simultaneously analyze. 
\editm{Previously, SPLOC \cite{grear2021molecular} was shown to successfully classify the classical iris and wine data sets and identify latent signals embedded within a noisy environment while mitigating false identification of noise as signal. The discriminate subspace could accurately reconstruct the latent signal without over-fitting to noise. When applied to the intricate problem of identifying functional dynamics hidden within the atomic trajectories of beta lactamase enzymes, each involving 789 df, SPLOC was able to pinpoint key features of atomic motions. Regarded as a working hypothesis, these identified collective motions are learned as being critical for specific mutants to function. Furthermore, an iterative learning process for discovery of new molecules was introduced, which is a self-directed learning process that involves the formation and refinement of a working hypothesis as new data is presented to the machine. This past work motivates the question of how biases alter the discovery process, such as preset perspectives that amount to weighting certain information as more or less important during data exploration.} 


\editb{The work here utilizes a synthetic dataset to highlight the underlying effects of biases in perceptrons which can be controlled by modifying the functional form of the data-driven adaptive rectifying functions used in SPLOC. The machine perception is expressed by the coordinate system that the data is represented in as a complete basis set that spans the state space. Through the use of projection pursuit, biases manifest during the partitioning of an orthonormal basis.}
With implementation details published previously \cite{grear2021molecular}, only the salient components of SPLOC relevant to this study are highlighted. Particular emphasis is placed on the importance of bias in relation to data-driven hypothesis refinement, including bias from preconceived perspectives.


\section{Materials and Methods}
\subsection{Data}
\begin{figure}[h]
\centering
\includegraphics[width=1.0\textwidth]{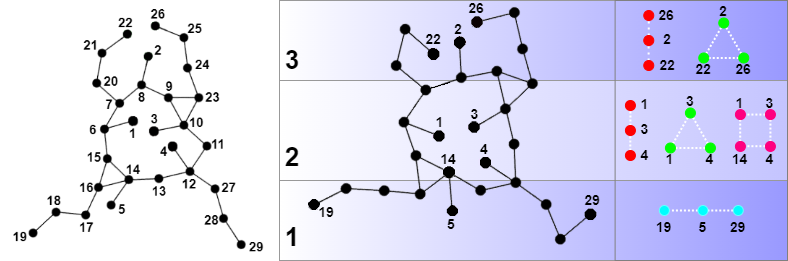}
\caption{(a) A synthetic molecule with atom labels. (b) Two partitions separate three domains labeled as 1, 2, 3. In each domain, the atoms can remain free or a subset of atoms can interact, forming geometrical signatures. The possible geometrical variants of each domain are shown in the far-right column.}
\label{fig:2dMol}
\end{figure}

The previously generated synthetic dataset of 24 two-dimensional molecules \cite{grear2021molecular} is reused for illustration. \editm{This dataset contains challenging spatio-temporal patterns to identify, yet it is simple enough for the results of the analysis to be clear and certain.}  Each molecule contains \editr{$N_a = 29$} atoms, resulting in \editr{$p = 58$} df. All trajectories are comprised of 20000 conformations. Each of these trajectories is divided in half, spawning two data streams, each having 10000 observations that yield 142.41 observations per variable. The molecules differ due to subtle embedded atomic interactions that create certain geometrical signatures. The nomenclature for each molecule is expressed as $abc$ to reference geometrical signatures within each domain, as depicted in Figure \ref{fig:2dMol}. 

When no underlying pattern over time exists within a given domain, this is annotated by F (free). The free conformation is the initial default for all molecular dynamics. The available geometrical signatures for each domain are: Domain 1: E (extended); Domain 2: L (linear), S (square), T (triangular); Domain 3: L (linear), T (triangular). The partial permutations results in 24 molecules exhibiting distinct dynamics. 

In this study, the set of 24 molecules is divided into two classes. The functional class includes the four molecules EFL, ELL, ETL and ESL, which are of the type E$b$L where $b$ can be F, L, T or S. These functional molecules are constrained to be in the extended (E) geometry in the first domain and have a linear (L) geometry in the third domain. There are no restrictions on the type of geometry that can be explored in the second domain. This designation mimics an allosteric interaction, in which ligand binding at one location, modeled as additional interactions shared among a subset of atoms, may impact the motions and ligand binding affinity at a distant location on the molecule. Among the 20 remaining nonfunctional molecules, three subgroups of the form F$b$F, $ab$F and F$bc$ are considered, respectively having 4, 8 and 12 molecules. Different scenarios of training compare 4 functional molecules of type E$b$L against the 4, 8, 12 molecule subgroups and all remaining 20 molecules. 

\editr{For each training scenario the goal is for SPLOC to learn the particular structural motif which differentiates the functional from nonfunctional molecules.}
In this work, SPLOC is run with five perspectives: -2, 0-, 0, 0+, and +2 \editr{to explore the impact that bias has on the}  \editm{relevance of information that} \editr{is explored} \editm{while forming a hypothesis}. Moreover, SPLOC was run 100 times for each training scenario/bias combination to characterize the variations in the solutions that SPLOC \editm{produces}.
\editr{The hypothesis SPLOC builds takes the form of a complete basis vector set in which the data can be represented to highlight similarities and differences. This orthonormal basis is learned through data exploration, where the hypothesis partitions the high dimensional data into discriminant features described by a subset of the basis vectors that span the discriminant subspace, and indifferent features described by a subset of basis vectors that span the indifferent subspace.}

\editm{Data streams are input to SPLOC as separate ensembles, each defining a data packet. A data packet}  \editr{is constructed from m distinct p dimensional observations of a system. For the synthetic data employed here, a single data sample is a conformation of a molecule in a simulation, represented by the state vector $\vec{x}(t)=[x_1, .. x_{N_a}, y_1, .., y_{N_a}]$. A data stream is an ensemble of conformations that sample the dynamics of the system, $X=\{\vec{x}(0), \vec{x}(1), \vec{x}(2), .., \vec{x}(m)\}$, as a series of frames or snapshots. A data packet is labeled functional or non-functional depending on whether it contains the geometrical signature of interest.}

\subsection{Projection pursuit machine learning}
Data \editm{packets} with $p$ df are described by a complete orthonormal basis set, where each basis vector (a.k.a. mode) defines a projection direction. 
\editr{Starting from either the natural data basis, a basis constructed from PCA, or any complete set of orthogonal vectors,} SPLOC optimizes the complete basis in $p$-dimensional space to elucidate similarities and differences between data \editm{packets}. \editr{In this work the starting basis for SPLOC was chosen to be constructed from PCA.}
\editr{For $p$ modes, a $2p$ dimensional feature space can be constructed from the emergent properties of the system. These properties are the mean, $\mu(m)$, and standard deviation, $\sigma(m)$, of the projections of each data packet along each of the basis vectors. SPLOC attempts to cluster the data in each two dimensional cross section of this feature space representing the $\mu(m)$ and $\sigma(m)$ of a particular mode $m$. This cross section is called a \textit{mode feature space plane} (MFSP) and is an essential component of the objective function used in SPLOC that is maximized.}

\editm{The objective function is net efficacy (E), which is linearly separable as the sum over the efficacy of each mode, $E(m)$.} 
Based on the consensus of emergent properties over all \editm{pairs of} data streams, each mode is evaluated for (1) {\it selection} power, $S(m)$, that quantifies signal-to-noise; (2) {\it consensus} power, $C(m)$, that quantifies statistical significance, and (3) {\it quality} \editm{of clustering within a MFSP. The conditional selection power,  $S(m| \alpha, \beta )$, associated with two data packets, $\alpha$ and $\beta$ respectively representing functional and nonfunctional classes, is calculated for mode $m$ using the formula:

\begin{equation}
    \displaystyle S(m | \alpha, \beta ) = \begin{cases}
    \sqrt{snr(m, \alpha, \beta )^2 + rex( m, \alpha, \beta )^2}+1 & \text{if $< S_i$} \\
    \sqrt{sbr(m, \alpha, \beta )^2 + rex( m, \alpha, \beta )^2}+1 & \text{if $> S_d$} \\
    S_o                                                           & \text{otherwise} 
    \end{cases}
\end{equation}

\noindent where $snr$ is the signal-to-noise-ratio given by 
$snr(m,\alpha,\beta) = |\mu(m | \alpha)-\mu(m | \beta)| / \sqrt{ \sigma(m | \alpha)^2 +\sigma (m | \beta)^2}$; $sbr$ is the signal-beyond-noise defined by
$sbr(m,\alpha,\beta) = \max \left[ 0, snr(m,\alpha,\beta)-1 \right]$, and $rex$ is the excess ratio of standard deviations defined as $rex(m,\alpha,\beta) = \max \left[ \frac{\sigma (m | \alpha)}{ \sigma (m | \beta) } , \frac{ \sigma ( m | \beta)}{ \sigma ( m | \alpha)} \right] - 1$. The values of $S_i$ and $S_d$ are respectively 1.3 and 2.0 for the upper threshold for indifference and the lower threshold for discrimination, and $S_o = 1.6125$ as their geometrical mean given as $S_o = \sqrt{S_i S_d}$}.

The collection of results for emergent properties of the data streams obtained for a specific mode, $m$, must be statistically consistent across all data stream pairs. The consensus measure $C(m)$ quantifies this consistency level \editm{using modified logistic functions as explained in REF \cite{grear2021molecular}.} If different data stream pairs between functional and nonfunctional classes cannot reach a statistically significant consensus on whether there exists a difference or similarity between the two classes, then the information associated with that mode is set as undetermined. In addition, within-class consistency between data streams is sought after \editm{as a geometrical property within each MFSP}, but this \editm{desired characteristic of data spread is} not reflected in the $C(m)$ measure. To reduce uncertainties in statistical estimates, various sampling methods can be applied when collecting data streams \editm{to increase the number of data packets to compare.} 

\begin{figure}[h]
    \centering
    \includegraphics[scale=0.456,trim=143 25 110 25]{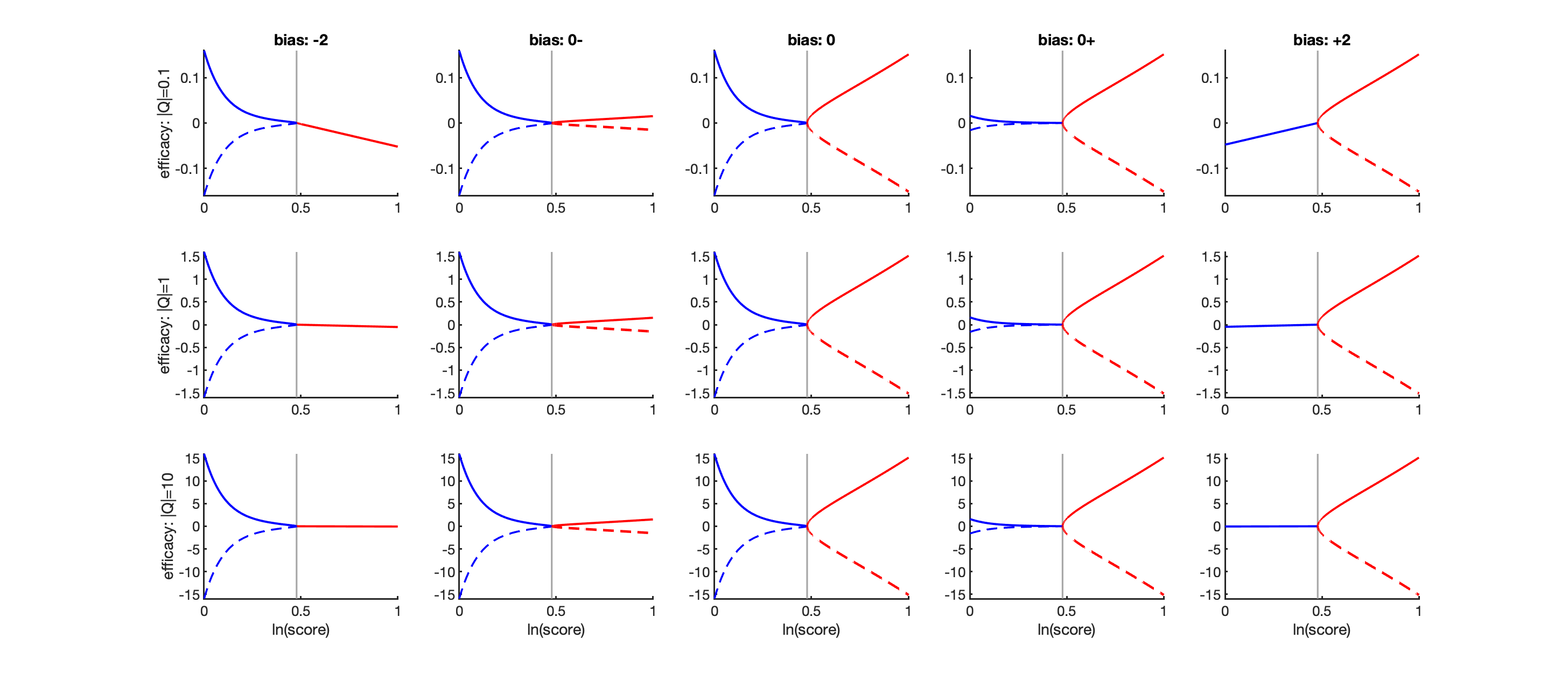}
    \caption{Functional forms for the rectifying adaptive nonlinear units under different biasing cases (columns) and intensity of MFSP clustering quality (rows). The light gray vertical line marks the reference score, $S_o$. Red and blue solid lines track the $r_d$ and $r_i$ functions, while dash lines indicate reverse biasing.}
    \label{fig:ranu}
\end{figure}

Letting $x = \log [S(m)/S_o]$, the efficacy per mode is given by:

\begin{equation}
\displaystyle E(m) = \begin{cases}
 Q_d(m) \times r_d(x) & \text{if $S(m) \ge S_o$ \hspace{5mm} } \\ 
 Q_i(m) \times r_i(x) & \text{if $S(m) < S_o$ \hspace{5mm} }  
 \end{cases}
 \label{eq:RANU}
\end{equation}

\noindent \editm{where $Q_d(m)$ and $Q_i(m)$ are geometrical clustering quality factors associated with the MFSP \cite{grear2021molecular}. For this work, it is only important to note that when data packets separate well across a decision boundary within the MFSP, $Q_d > 0 $ while $Q_i < 0$. Alternatively, when the functional and nonfunctional points in the MFSP mix well, then $Q_d < 0$ while $Q_i > 0$. These quality factors quantify how well the emergent properties spatially cluster within a MFSP, and they are completely data-driven based on observation. Furthermore,} the functions $r_d(x)$ and $r_i(x)$ control the biasing, and these functions are visualized in Figure \ref{fig:ranu}.

\editr{The algorithm for learning an orthogonal basis set in SPLOC is presented for completeness in Algorithm 1.}

\begin{figure}[h]
\centering
\includegraphics[width=.85\textwidth]{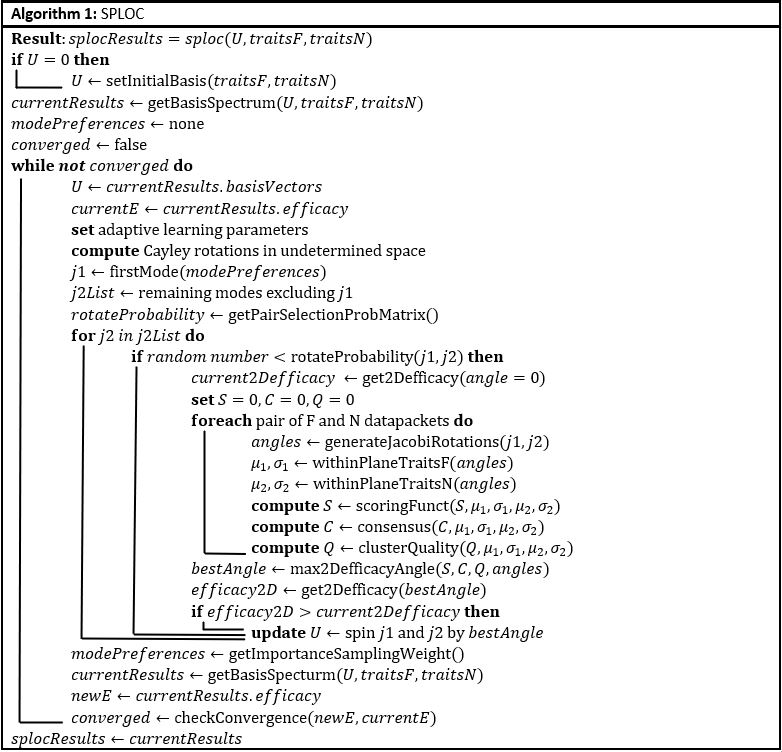}
\end{figure}


Mode efficacy increases when the mode can either resolve differences or similarities between functional and nonfunctional data streams, which respectively occur when $S(m) \ge S_o$ or $S(m) < S_o$. Modes that discriminate between functional and nonfunctional data streams are called d-modes; conversely, modes that quantify similarity through indifference are called i-modes. As orthogonal rotations of the basis vectors are performed, the efficacy of some modes may increase at the expense of others, \editm{creating} a multi-objective optimization problem \cite{roostapour2022pareto} \cite{jin2008pareto}\cite{ngatchouSystems}. As optimization proceeds, $S(m)$ is bifurcated away from $S_o$ while consensus and MFSP cluster quality are improved in relevant emergent properties. The collection of all modes defines an orthonormal basis which grounds the perception of a multivariate hypothesis. The consistency of a proposed hypothesis across the data is evaluated using the decision triad. This triad requires simultaneously exceeding minimum thresholds for selection power (triad member 1), consensus power (triad member 2) and MFSP cluster quality (triad member 3) in order to declare a mode as either a d-mode or i-mode; otherwise, the mode is considered undetermined (u-mode). 

The process of collapsing high-dimensional data onto a line for a mode projection represents a tremendous loss of information in exchange for an immense gain in specificity. Although no information is lost when using a complete orthonormal basis, how information is distributed is not unique. Each mode is mapped to a perceptron which is governed by a rectifying adaptive nonlinear unit (RANU). For the $m$-th mode, when $S(m) > S_o$ this forms a data-driven hypothesis that there is a difference between functional and nonfunctional data streams. Conditional upon a statistically significant consensus, the proposed hypothesis is confirmed when the quality of clustering within the MFSP for a d-mode is positive \editm{($Q_d > 0$)}. Upon confirmation, mode efficacy increases through Eq. \ref{eq:RANU} to create a stronger belief supported by the observed projected data. Conversely, if the quality of clustering within the MFSP for a d-mode is negative \editm{($Q_d < 0$)}, the proposed hypothesis is rejected, and mode efficacy decreases through Eq. \ref{eq:RANU} using a reverse bias. An analogous process is applied to i-modes. The multi-objective optimization is then carried out simultaneously over all modes through competitive learning, where the recurrent NN will have an abundance of diverse perspectives across the RANUs during data exploration, as depicted in Figure \ref{fig:ranu}. 

The proportion of modes that describe how similar or different data streams appear depends on the perspective of finding differences or similarities more important. To bias importance, relative weights in a RANU are adjusted. For predisposed perspectives, reverse biases are introduced to reduce efficacy when unwanted results are found. In this work, five bias-cases are considered: predisposed bias toward i-modes (-2), weak bias toward i-modes (0-), unbiased (0), weak bias toward d-modes (0+) and predisposed bias toward d-modes (+2). The 0- bias sets $r_d(x) \rightarrow \frac{1}{10} r_d(x)$ and the 0+ bias sets $r_i(x) \rightarrow \frac{1}{10} r_i(x)$. A bias of (-2, +2) introduces an indiscriminate fixed level of reverse bias when a (d-mode, i-mode) is found since a predisposed perspective is not data-driven. It is worth noting that bias cases of -1 and +1 exist in SPLOC. These biases are not considered here because they are adaptive. The -1 and +1 cases start like their 0- and 0+ counterparts. However, as more (i-modes, d-modes) are discovered, the scale factor relaxes from $\frac{1}{10}$ to 1, corresponding to the unbiased case.

\subsection{Subspace Comparison}
The projection spaces obtained by SPLOC were compared by computing the mean square inner product (MSIP) between basis sets. The MSIP between two subspaces $U$ and $V$ spanned by vectors $\vec{u}$ and $\vec{v}$ is given by 

\begin{equation}
\mbox{MSIP}(U,V) = \frac{1}{\max[\dim(U), \dim(V)]}  \sum_{\vec{u} \in U} \sum_{\vec{v} \in V} (\vec{u} \cdot \vec{v})^2.
\label{eq:rmsip}
\end{equation}


\section{Results}
\begin{figure}[h]
    \centering
    \includegraphics[width=1.0\textwidth]{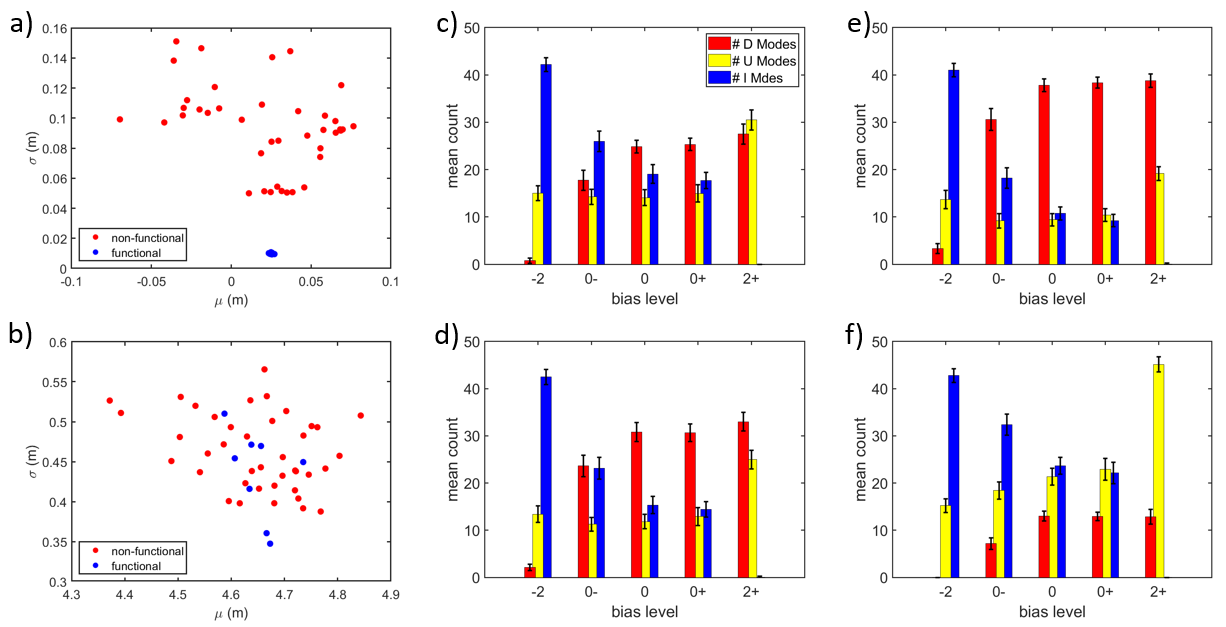}
    \caption{(a) \editm{MFSP for an} example d-mode (b) \editm{MFSP for an} example i-mode (c) E\textit{b}L vs F\textit{bc} modes (d) E\textit{b}L vs \textit{ab}F modes (e) E\textit{b}L vs F\textit{b}F modes (f) E\textit{b}L vs All modes. Shown for each training scenario are the averages from 100 replicates for extracted {D, U, I} modes with standard error bars. \editb{The figure legend of (c) applies to (d), (e), and (f).}}
    \label{fig:modes}
\end{figure}

In Figure \ref{fig:modes}, panels (a) and (b) show a typical MFSP for a d-mode and i-mode. These MFSPs will yield a positive and negative cluster quality when the selection power for the mode is greater than the reference value, $S_o$. However, the same MFSPs will yield a negative and positive cluster quality when the selection power for the mode is less than $S_o$. The sign of the cluster quality is a bias based on observation. If selection power suggests a projection direction will discriminate between functional and nonfunctional molecules, and the MFSP shows poor clustering, then the proposed direction is a mistake, and alternative projections will be sought. Otherwise, good clustering confirms the projection direction is correct, and after some fine tuning the d-mode will be locked in place once it cannot increase efficacy at the expense of any other mode. 
  
For each bias-case, the average numbers of d-modes, u-modes and i-modes are shown in Figure \ref{fig:modes} (c-f). The number of i-modes forming the indifference subspace is strongly dependent on the bias, while the number of d-modes forming the discriminant subspace is dramatically affected only for the extreme bias toward i-modes by penalizing d-modes.  Importantly, the number of d-modes found with 0-, 0, 0+ and +2 bias was approximately the same. This indicates that SPLOC does not overestimate the discriminate subspace in the presence of a large statistical sample size. In contrast, more similarity can be found using a perspective where differences are considered less important. 

\begin{figure}[h]
    \centering
    \includegraphics[width=1.0\textwidth]{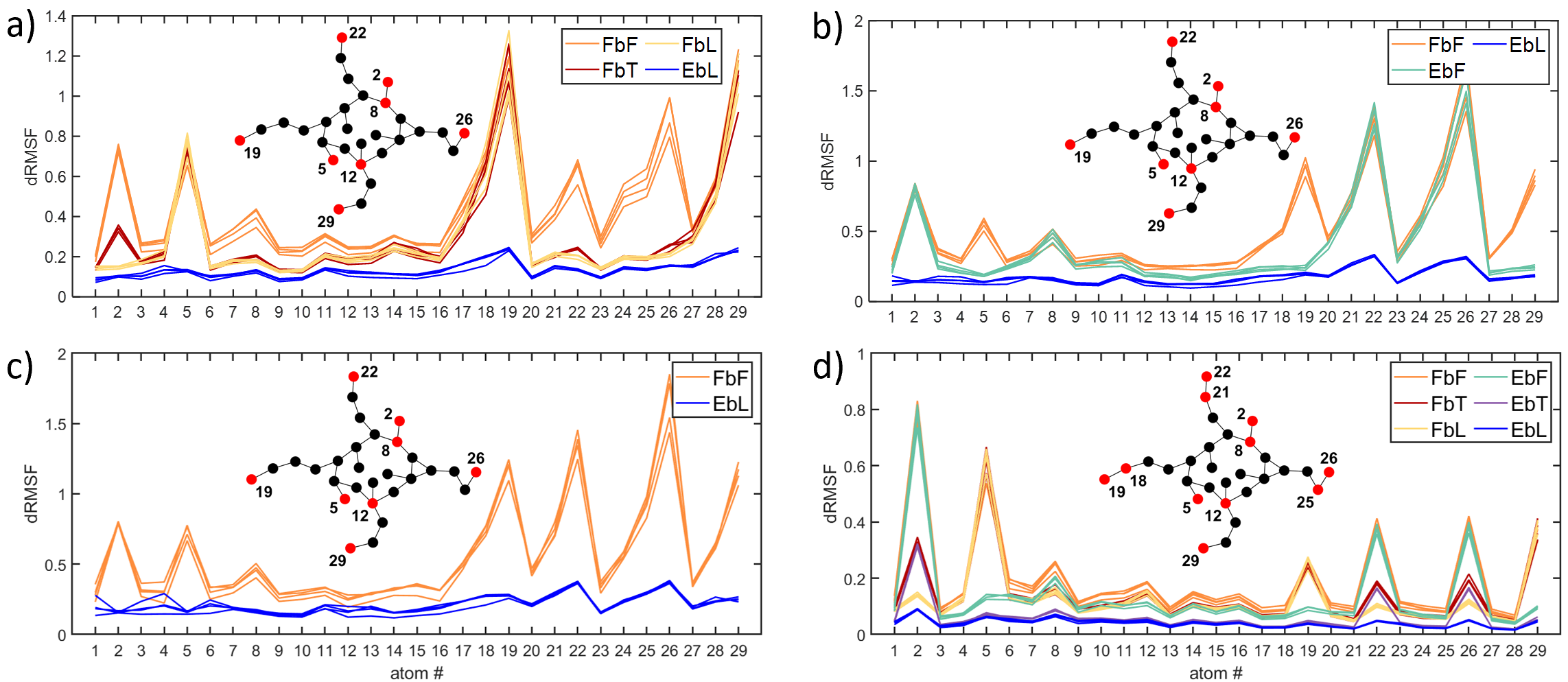}
    \caption{The dRMSF of each molecule is shown for the unbiased case under four training scenarios. The functional molecules included  E\textit{b}L, while the nonfunctional molecules included: (a) F\textit{bc} (b) \textit{ab}F (c)  F\textit{b}F (d) all others. Different molecular types with similar dRMSF are grouped by color in each panel. There are several clear differences in dRMSF between functional and nonfunctional groups at certain atoms. The inset highlights these atoms in red.}
    \label{fig:rmsf}
\end{figure}
        
The dynamical motions of the molecules are readily interpretable through the discriminant and indifference subspaces, where the emergent properties of a MFSP show which molecules are similar or different. The d-modes and i-modes are used to project the original data into the discriminant or indifference subspaces. The concept of functional dynamics is meaningful as the dynamical motions projected into the discriminate subspace identifies likely functional mechanisms, whereas the extracted features from most other types of NN are usually obfuscated. 

A common way to characterize atomic motion is through the root mean square fluctuations (RMSF). When atomic motions are first projected with d-modes or i-modes, a dRMSF or iRMSF is constructed. Information contained in dRMSF and iRMSF is readily mapped to specific atoms, which helps interpret mechanisms responsible for functional dynamics. When comparing iRMSF across all molecules, the indifference subspace identified conserved motions across all molecules as expected. However, unexpectedly the iRMSF were similar across all training scenarios (data not shown). More interestingly, dRMSF highlights differences between functional and nonfunctional molecules shown in Figure \ref{fig:rmsf}. 

The differences in dRMSF depend on training scenarios. When a group of similar nonfunctional molecules are compared to the functional molecules, many differences are uncovered, but only a subset of these differences may be responsible for functional dynamics. Over the four training scenarios, as more diverse nonfunctional molecules were included in the training scenario, the number of d-modes decreased because the essential elements of function are being uncovered. An over zealous working hypothesis will be associated with a larger discriminant subspace, as some differences are irrelevant to function. As reported previously \cite{grear2021molecular}, the working hypothesis narrows as more constraints are introduced as diverse molecules are contrasted. This is a consequence of an inductive reasoning where a hypothesis is proposed based on the consistency of examples observed, and when contradicting data is found, a revision of the hypothesis is often necessary.  

Interpreting data and hypothesis building in PP is exemplified in figure \ref{fig:rmsf} (a), where E\textit{b}L is compared with F\textit{bc}. Major differences in dRMSF are found at atoms \{2,5,19,22,26,29\}. Within functional molecules, atoms \{5,19,29\} make up the extended signature in domain 1, and atoms \{2,22,26\} make up the linear signature in domain 3. By construction, these two domains are precisely where differences between the functional and nonfunctional molecules should be found for optimal discrimination. The dRMSF for all nonfunctional molecules at domain 1 atoms show strong similarities to each other, and differences to the functional molecules, which indicates a conserved discriminating feature. The differences at domain 3 are not universal as demonstrated by the segmented dRMSF in nonfunctional molecules. The F\textit{b}L and F\textit{b}T molecules show more similar dRMSF to the functional class because they share constraints in this domain, whereas F\textit{b}F molecules show much higher fluctuation because there are no extra constraints. In this case SPLOC learned that the main differentiating feature between functional and nonfunctional molecules is the F versus E conformation in domain 1. An additional hypothesis that reflects differences between L or T versus F conformations in domain 3 was also proposed.

\begin{figure}[h]
    \centering
    \includegraphics[width=1.0\textwidth]{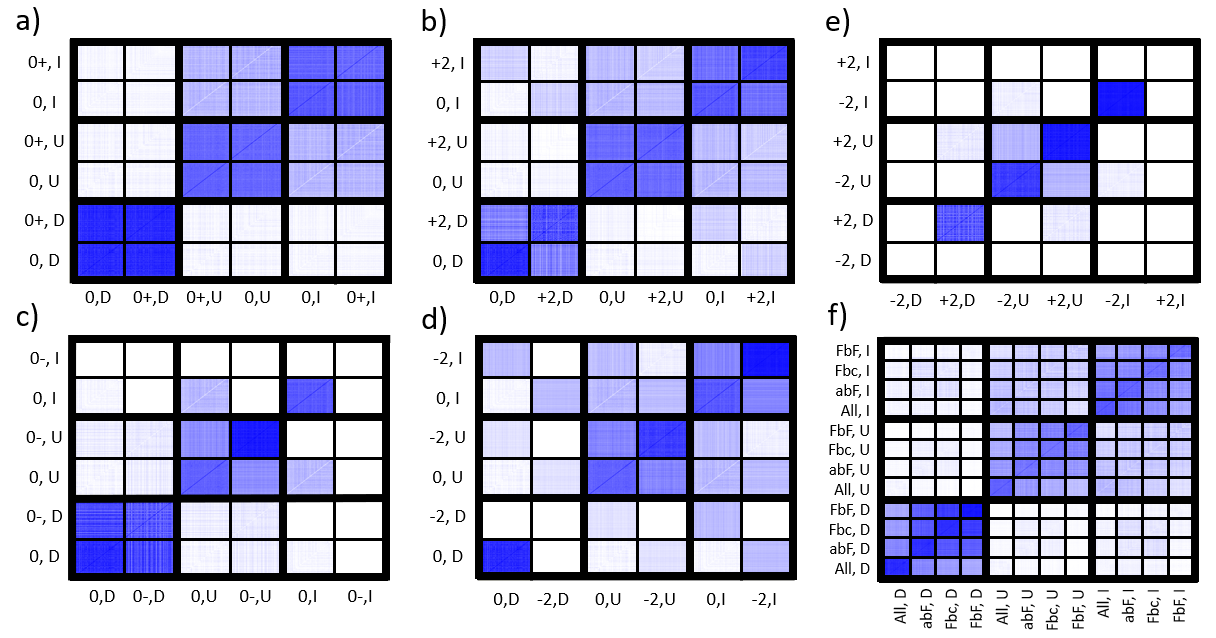}
    \caption{A collection of MSIP plots show subspace comparisons between SPLOC solutions obtained using different bias modes and training scenarios. In panels a-e, all plots were obtained using E$b$L as functional and all other molecules as nonfunctional. (a) unbiased vs weakly positively biased (b) unbiased vs strongly positively biased (c) unbiased vs weakly negatively biased (d) unbiased vs strongly negatively biased (e) strongly negatively biased vs strongly positively biased. In (f) a comparison between the projection spaces of all training models is shown for the unbiased case.}
    \label{fig:rmsip}
\end{figure}

By keeping a complete set of basis vectors, information is never lost regardless of bias, simply interpreted differently by the model via perspective.  MSIP was used to compare the similarity in SPLOC solutions. In figure \ref{fig:rmsip}  (a-d) the unbiased SPLOC mode is compared with the various biased subspace results for D, U, and I modes. Figure \ref{fig:rmsip} (a) and (c) show that information in promiscuous u-modes can be rotated into and out of the D an I spaces, with a degree of difficulty depending on the bias. Specifically with bias 0+, information is exchanged between the U and I space and with bias 0- there is a weak exchange of information between the D and U spaces. This is due to the change in perspective to favor discriminant or indifferent features. Comparing extreme biases with the unbiased case in figure \ref{fig:rmsip} (b) and (d), all relevant information is lumped into the discriminant or indifferent subspace, depending on the extreme. 

Significant d-modes can be extracted from the U space, but when aggressively searching for i-modes, information that was discriminant can be forced into I space, as shown in figure \ref{fig:rmsip} (d). This clearly demonstrates that based on perspective, data that looks different under one bias now looks the same under another bias. It is worth pointing out that the MSIP analysis presented here is not saying the entire mode switches from one subspace to another, but that various linear combinations of modes from one space switches out from one space and into another. These type of alternative views will increase as the number of df increases. When comparing the biases -2 and +2, \ref{fig:rmsip} (e), little information is shared between D and I modes, as the algorithm aggressively tries to fill either the D space or the I space, however, their U spaces are quite similar. The u-modes with low statistical support may facilitate alternate hypotheses across nonfunctional molecules that behave differently from one another.

Figure \ref{fig:rmsip} (f), compares the SPLOC solutions across the different experiments. The D spaces for F\textit{bc}, \textit{ab}F, or F\textit{b}F training sets were very consistent. When all molecules other than E\textit{b}L were considered nonfunctional, the decrease in number of d-modes due to SPLOCs refined hypothesis of functionality led to a decrease in D space MSIP with the other training sets. The D space obtained with all biases except for (-2) captured the functional dynamics faithfully needed for accurate classification.

\section{Conclusion}
For the synthetic data set analyzed here, SPLOC successfully identified the major differences between functional and nonfunctional molecules, and found that the shared conserved properties across all molecules are markedly high. For extreme preconceived perspectives that either discount the possibility of finding differences or similarities, SPLOC could neither confirm or deny such a belief, because data supporting the contrary-view was not taken into account due to the predisposed bias. However, this latent information is retained in the u-modes that can be extracted to refine the working hypothesis once supporting data for the contrary view is taken into account. \editm{Outside of these extreme cases, in general the biasing for refining a working hypothesis depends on whether more data is found to support the proposed claim either through augmenting training data with more consistent examples or by optimizing the vantage point as basis vectors rotate. In SPLOC, the recurrent NN is comprised of a heterogenous set of perceptrons, where each perceptron has a custom rectifying unit that adapts to the data that is observed through exploration. Mistakes are found through inconsistencies, where parts of the multivariate hypothesis are rejected, which leads to refinement.} These results suggest that an abundance of perspectives across perceptions comprising a NN is a good strategy to achieve robust solutions with maximum consistency for an objective analysis that minimizes preconceived prejudices with minimal risk of missing latent features. 

All relevant code to SPLOC can be found at \url{https://github.com/BioMolecularPhysicsGroup-UNCC/MachineLearning/tree/master/SPLOC}. The data set for the toy molecules can be downloaded in full at \url{https://zenodo.org/record/4465089#.YaovumDMKl5}.

\section{Conflicts of Interest}
The authors acknowledged no conflict of interest. 
\section{Acknowledgments}
We acknowledge the provision of additional computing facilities and resources by the University Research Computing group in the Office of OneIT at the University of North Carolina at Charlotte.

\bibliographystyle{IEEEtran}
\bibliography{main.bib}

\end{document}